\documentclass[12pt, letterpaper]{article}

\usepackage[parfill]{parskip}
\usepackage{graphicx}
\usepackage{float}
\usepackage{times}
\usepackage[top=1in, bottom=1in, left=1.25in, right=1.25in]{geometry}
\usepackage[stretch=20,shrink=50]{microtype}

\begin{document}\pagestyle{plain}

\title{Automated Assignment of Backbone NMR Data using Artificial Intelligence}

\author{John Emmons$^*$, Steven Johnson$^*$, Timothy Urness$^*$ \& Adina Kilpatrick$^+$\\\\
  $^*$Department of Mathematics and Computer Science\\
  $^+$Department of Physics and Astronomy\\
  Drake University\\
  Des Moines, IA\\
}

\date{} 

\maketitle

\abstract{Nuclear magnetic resonance (NMR) spectroscopy is a powerful method for
  the investigation of three-dimensional structures of biological molecules such
  as proteins. Determining a protein structure is essential for understanding
  its function and alterations in function which lead to disease. One of the
  major challenges of the post-genomic era is to obtain structural and
  functional information on the many unknown proteins encoded by thousands of
  newly identified genes. The goal of this research is to design an algorithm
  capable of automating the analysis of backbone protein NMR data by
  implementing AI strategies such as greedy and A* search.}

\thispagestyle{empty} \pagebreak

\section{Nuclear Magnetic Resonance (NMR)}
Nuclear magnetic resonance is a phenomenon in which atomic nuclei absorb
electromagnetic radiation at frequencies related to their chemical properties
and the local molecular environment. Biophysicists use this property to gain
structural knowledge of biomolecules, including proteins, DNA and RNA. NMR
spectroscopy is currently the only method that allows the determination of
atomic-level structures of large biomolecules in aqueous solutions similar to
their in vivo physiological environments.

\medskip

Several types of NMR experiments can be used in the analysis of protein
structures. In particular, essential information is provided by the chemical
shifts of NMR-active nuclei present in proteins, including hydrogen and isotopes
of carbon and nitrogen. The chemical shift is a quantifier for the deviation in
the resonant frequency of a nucleus from its value in a structure-free
environment, and therefore provides information on the local conformation.
Determining the chemical shifts of all or most of the nuclei in a biomolecule is
the first step in determining its structure.

\subsection{NMR Assignment Methodology}

\begin{figure}[h]\centering
  \includegraphics[width=0.5\textwidth]{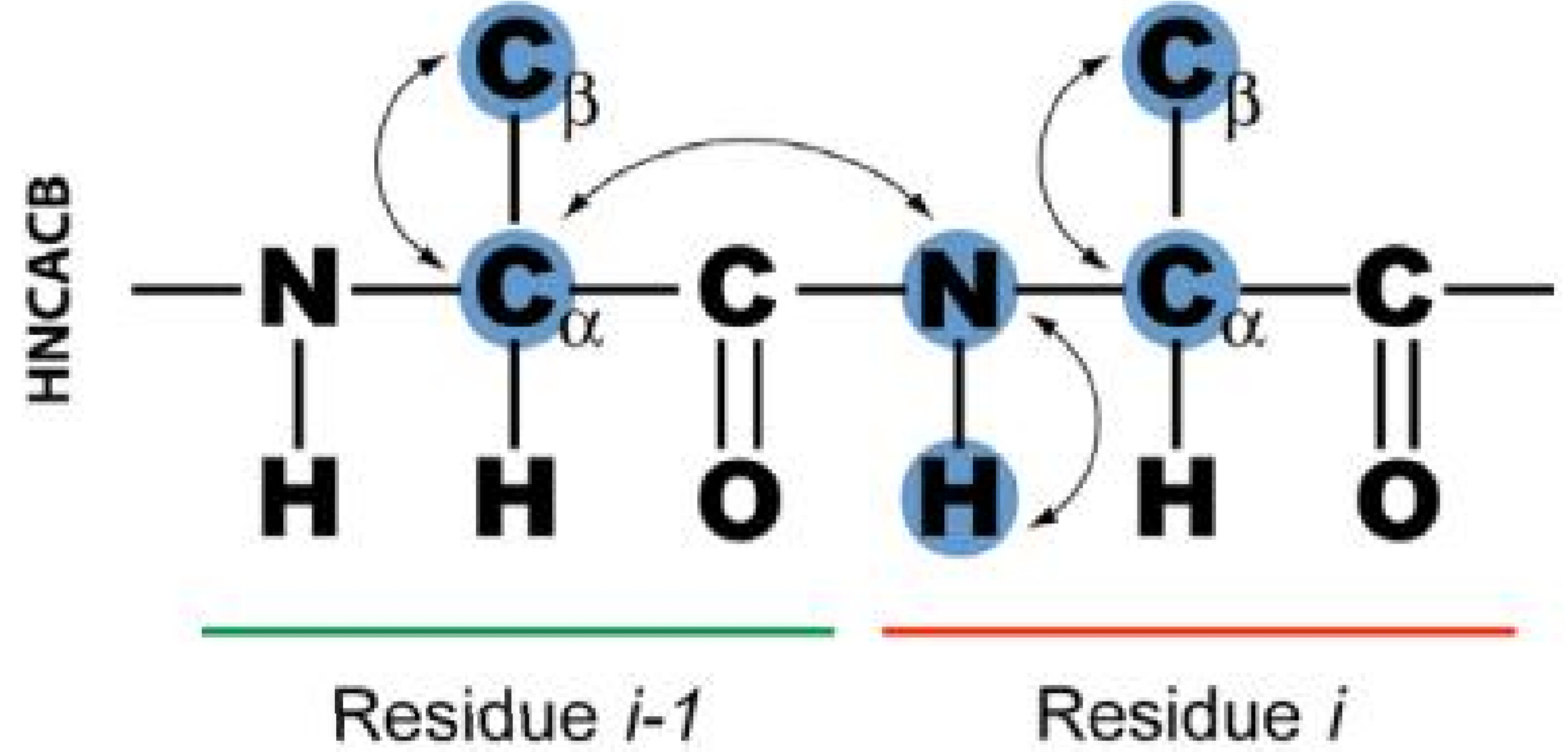}
  \caption{HNCACB NMR experiment.}\label{fig:hncacb}
\end{figure}

An important set of chemical shifts in a protein are those corresponding to the
backbone nuclei, including the nitrogen, attached hydrogen, and the alpha and
beta carbon atoms (C$\alpha$ and C$\beta$) of each of the residues. The backbone
residues constitute the building blocks of a protein chain (Figure
\ref{fig:hncacb}). These checmical shift signals are measured using various NMR
experiments, and then matched to the individual residues in the protein in a
process called sequential assignment.

\medskip

A prerequisite of the assignment process is data collection using experiments
that can provide information about the connectivities between neighboring
residues \cite{1}. One such experiment, called HNCACB, yields signals
corresponding to the C$\alpha$ and C$\beta$ nuclei of one residue in the protein
(residue $i$) , plus the C$\alpha$ and C$\beta$ signals of the immediately
preceding residue (residue $i - 1$) (Figure \ref{fig:hncacb}). A second
experiment, CBCA(CO) NH, can be used to yield the chemical shifts of the
preceding residue only. These experiment are not independent, so they allow
scientists to distinguish unambiguously between signals from residue $i$ and
residue $i - 1$.

\medskip

\begin{figure}[t]\centering
  \includegraphics[width=0.5\textwidth]{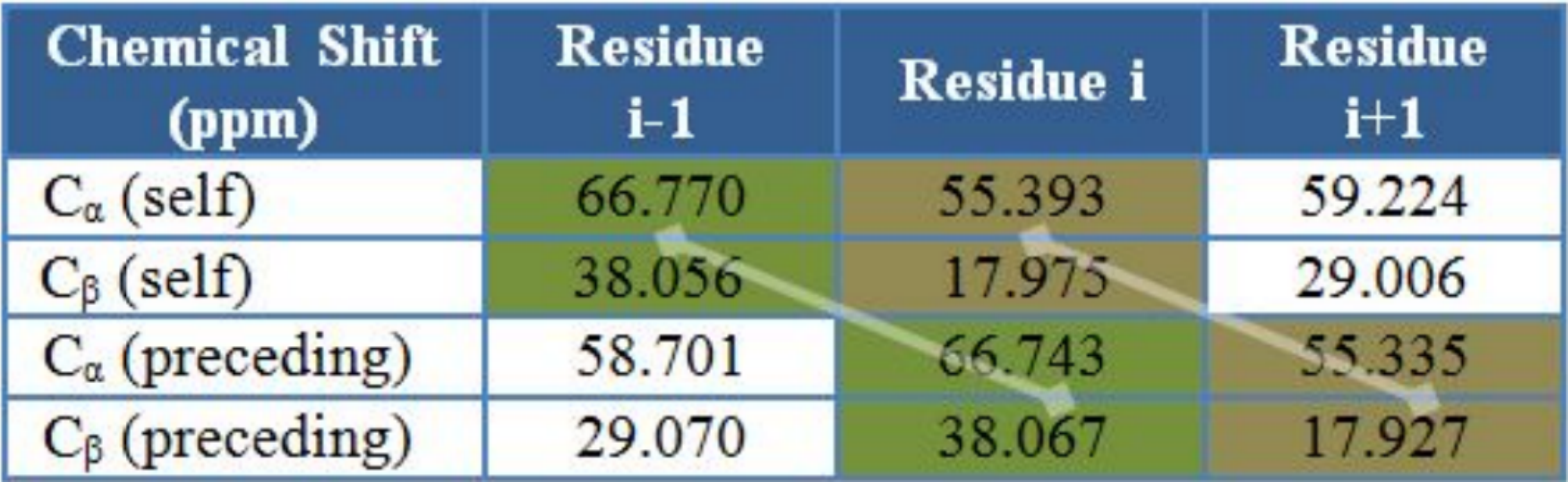}
  \caption{Sequentially matched backbone carbon signals from HNCACB chemical shifts.}\label{fig:sequentialMatching}
\end{figure}

Analysis of all inter-residue connectivities allows the linking of signals from
each backbone atom with signals from their preceding neighbor, creating a
pattern of sequentially linked chemical shift values which reflects the
sequential linear arrangement of the individual residues in the protein sequence
(Figure \ref{fig:sequentialMatching}). This pattern is then matched with the
protein sequence, by using the fact that certain residues have characteristic
C$\alpha$ and C$\beta$ chemical shift ranges which uniquely identify them. Thus,
each measured chemical shift is assigned to a location in the protein, and this
information can then be used to infer structural information about the
biomolecule.

\section{Manual Procedure}

The sequential assignment of backbone NMR data is typically done
manually. However, the process is very time-consuming (manual assignment of NMR
datasets can take days to months) and is error-prone \cite{2}. Common
difficulties in manual data assignment arise from missing or ambiguous data, as
well as spectroscopy artifacts. Therefore, in many cases the data analysis
procedure is slow and nontrivial

\medskip

\begin{figure}[b]\centering
  \includegraphics[width=0.5\textwidth]{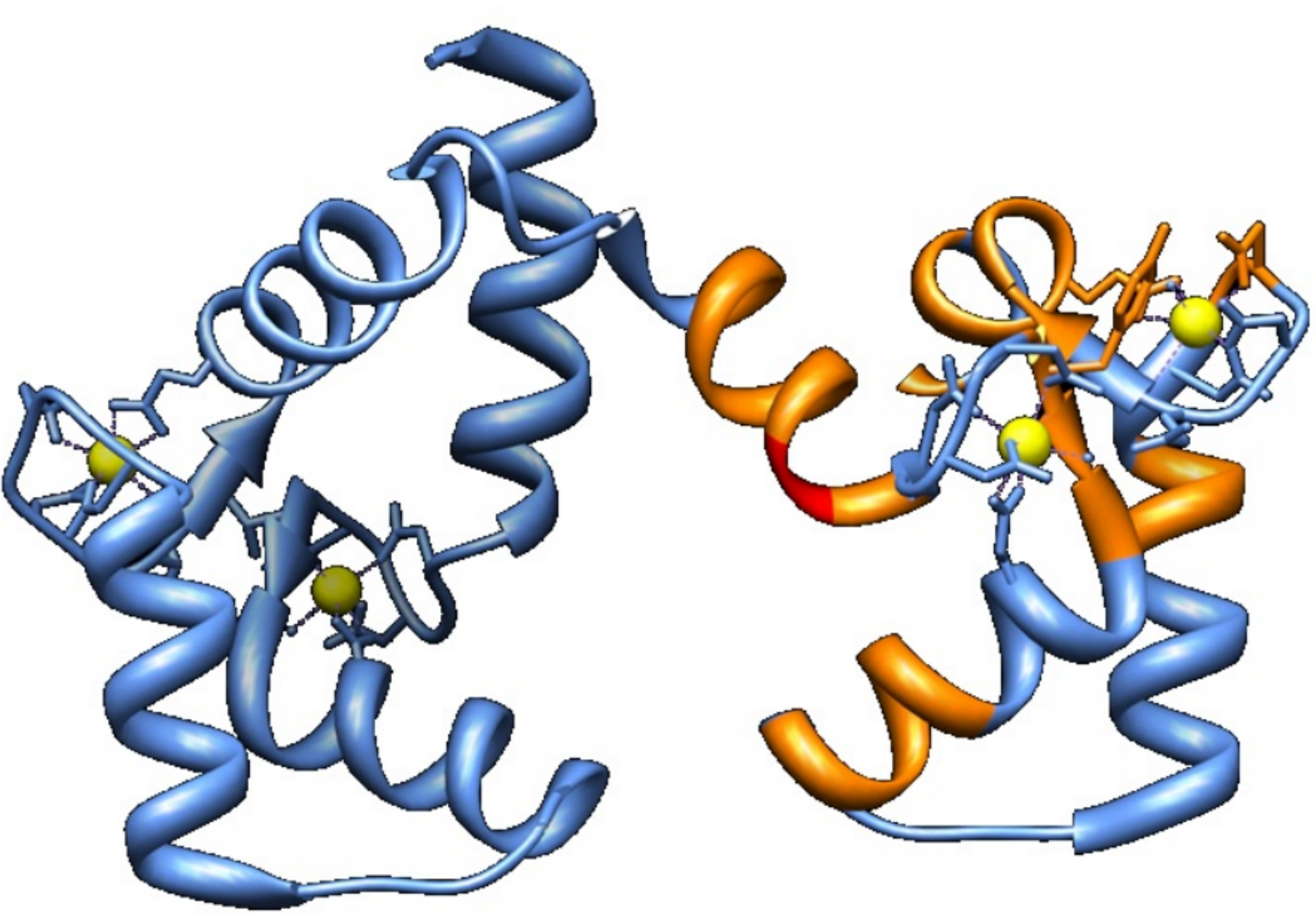}
  \caption{V91G calmodulin mutant investigated during the summer of 2012.}\label{fig:v91gCaM}
\end{figure}

One example is a chemical shift dataset acquired as part of a related research
project during the summer of 2012 (Figure \ref{fig:v91gCaM}). A sample of the
calcium-binding protein calmodulin, with a point mutation at position 91 in the
sequence, produced spectra with many missing and ambiguous data that required
several weeks to assign. This experience inspired our team to develop an
automatic assignment procedure.

\section{Automated Assignment}

Several different approaches to creating algorithms capable of automated
assignment of NMR datasets have been tried; nonetheless, there is still room for
improvement in the field. Since last fall, we have been developing an algorithm
capable of rapid, accurate assignment of small, yet non-trivial backbone NMR
datasets, using techniques from artificial intelligence and statistics.

\medskip

The current iteration of the algorithm assigns data in a three-step process: (1)
the protein sequence is searched for short stretches (subsets) containing easily
identifiable residues (those with high or low C and C chemical shifts); (2)
these subsets are matched with sequentially linked chemical shifts; (3) an
iterative greedy search algorithm completes the assignment, putting all residues
in a sequential order consistent with the protein sequence.  Steps (1) and (2)
reduce data complexity before computationally expensive methods are used in step
(3).

\medskip

In step (1), the protein sequence is examined for subsets of the full-length
sequence containing residue(s) with highly unique chemical shifts. The algorithm
will record a subset if it contains one or several adjacent highly unique
residue(s).

\medskip

In step (2), the recorded subsets are matched with chemical shifts corresponding
to their residue type(s). Statistical methods are used to ensure these chemical
shifts are correctly linked in an $i$ to $i - 1$ pattern. This process is done
iteratively, starting with a very low error tolerance which is gradually
increased until all recorded subsets are matched with sets of chemical
shifts. This produces sequentially linked residues that are subsets of the
full-length protein sequence. Each subset can be abstracted and treated as a
single pseudoresidue with C$\alpha$ and C$\beta$ of $i$ and $i – 1$
corresponding to the chemical shifts of the residues making up the front and
back of the subset, respectively.

\medskip

In step (3), all chemical shifts corresponding to pseudoresidues and the
remaining individual residues are placed in sequential order using an iterative
greedy search algorithm. In this method, an arbitrary starting residue or
pseudoresidue is first chosen. All remaining residues and psuedoresidues are
added behind this starting point, such that the residue or psuedoresidue
producing the lowest error in its C$\alpha$ and C$\beta$ $i - 1$ chemical shifts
is always placed next. This process is repeated until all residues and
psuedoresidues have acted as the starting point. The generated sequence that
produces the lowest total error (the sum of the errors between adjacent
residues) is chosen as the correct assignment.

\section{Preliminary Results}

Assignment of small test datasets has proven that this process can correctly
analyze NMR data; however, much room for improvement still exists. Research is
expected to continue in the spring of 2013 to verify that the algorithm can
correctly assign larger datasets, by incorporating aspects of machine learning.

\section{Future Research}

The accurate assignment of nontrivial NMR datasets is necessary for sustained
advancement in the fields of structural biology and proteomics. As NMR
technology advances, structural studies of very large molecules will become
possible, with the price of increased complexity in data assignment. This
research directly confronts the issue of nontrivial NMR datasets and is also
consistent with the interests of our group as students and faculty of computer
science, physics, and mathematics.

\end{document}